# AN ITERATIVE CLOSEST POINT METHOD FOR MEASURING THE LEVEL OF SIMILARITY OF 3D LOG SCANS IN WOOD INDUSTRY

Cyrine Selma[1,2], Hind Bril El Haouzi[1,2], Philippe Thomas[1,2], Jonathan Gaudreault[3,4], Michael Morin[4,5]

[1] Université de Lorraine, Vandœuvre-Lès-Nancy Cedex, France
[2] CNRS, CRAN, UMR7039, France. cyrine.selma.tn@ieee.org,
{hind.el-haouzi, philippe.thomas}@univ-lorraine.fr
[3] Department of Computer Science and Software Engineering, Université Laval, Québec, QC, Canada.
[4] FORAC Research Consortium, Université Laval, Québec, Canada.
jonathan.gaudreault@ift.ulaval.ca
[5] Department of Mechanical and Industrial Engineering, University of Toronto, Toronto, ON, Canada. mmorin@mie.utoronto.ca

**ABSTRACT**

In the Canadian's lumber industry, simulators are used to predict the lumbers resulting from the sawing of a log at a given sawmill. Giving a log or several logs' 3D scans as input, simulators perform a real-time job to predict the lumbers. These simulators, however, tend to be slow at processing large volume of wood. We thus explore an alternative approximation techniques based on the Iterative Closest Point (ICP) algorithm to identify the already processed log to which an unseen log resembles the most. The main benefit of the ICP approach is that it can easily handle 3D scans with a variable number of points. We compare this ICP-based nearest neighbor predictor, to predictors built using machine learning algorithms such as the K-nearest-neighbor (kNN) and Random Forest (RF). The implemented ICP-based predictor enabled us to identify key points in using the 3D scans directly for distance calculation. The long-term goal of this on-going research is to integrated ICP distance calculations and machine learning.

## 1. INTRODUCTION

The wood-products industry is facing numerous challenges regarding the maximization of profits and sales when sawmilling. Natural forest resources are characterized by their high heterogeneity and the sawing of a log at a given sawmill is a complex physical process of which the output is not easily forecasted. It is nonetheless critical to decide which contracts to accept based on raw material resources, which sawmill to supply from which cutblocks, and how to configure the parameters of the sawmilling equipment in order to improve the profits and the sales.

Sawing simulators, such as Optitek [1], have been used for years to forecast the production of sawmills and to help at these tasks. They are, for instance, especially useful for sawmill design where time is not an issue [5]. Given the three-dimensional (3D) scan of a log and a sawmill's model, a sawing simulator outputs, among other information, the different pieces of wood that will result from the sawing [1]. Unfortunately, simulators tend to be slow and complex to use when working on large customers' demand under time constraints.

Machine learning algorithms such as K-nearest neighbors (k-NN) [2], decision tree (DT) [3] and random forest (RF) [4] were proposed as a complement to sawing simulation [5]. The authors evaluate machine learning approaches at the task of predicting the lumbers produced at a given sawmill when processing a given log. Based on simple decision rules and using six characteristics of an input log instead of the 3D scan, the predictors built using a machine learning approach are able to approximate the simulator's response of a sawmill producing 19 different lumber products. The technique involved learning the relationship between an input log, described in terms of six features (volume, length, wide-end diameter, narrow-end diameter, and shrinking), and the quantity of every lumber type produced at the plant when sawing it [5].

In this work, we study the usability of the Iterative Closest Point (ICP) algorithm [7, 8] as a mean to measure the resemblance between an input logs and already sawn logs. The ICP algorithm has the ability to exploit the entire amount of data (3D scans) available to determine the level of resemblance between two logs by measuring the minimum distance between the two point clouds of their corresponding scans. We used a simple nearest neighbor approach where the produced basket is determined by calculating the minimums of the distances between the input log and the logs of a given training set. Although standard machine learning approaches outperform this simple predictive technique, this research shed some light on the principal noise sources that render the ICP approach less effective for prediction.

This paper is structured as follows. We overview the current approaches used to approximate the sawing process and discuss the possibility of an ICP-based predictor in Section 2. Then, in section 3, we explain the ICP method we retained. Experiments and results are presented in Section 4. We conclude in Section 5.

## 2. FROM SAWING SIMULATION AND MACHINE LEARNING TO ITERATIVE CLOSEST POINT

In North America, wood products are standardized by the National Lumber Grades Authority (NLGA) [6]. Each lumber type has specific dimensions, grade, value and price. Lumber production is characterized by a divergent co-productive system; for a single input, it simultaneously produces multiple outputs showing different characteristics. The sawing equipment selects the cutting pattern that maximizes the expected profit. As it is a commodity market, we expect to sell the entire production. However, not every mill will produce the same products from the same log. Knowing what each mill would produce from each log (or batch of logs) allows a company to make better decisions thus increasing its efficiency. Sawing simulators has been built especially for that purpose.

Taking the 3D scan of a log as inputs (Figure 1), a sawing simulator virtually processes the log and allows predicting the products that could be obtained at the modeled sawmill. However, when there are tons of logs that must be processed by sawing simulators in a short period, these simulators tend to be slow and complex to use.

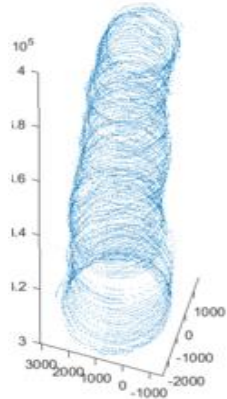

*Figure 1: Input of sawing simulators: 3D scan of a log*

Machine learning was recently proposed as a complement to simulation to overcome this difficulty [5]. The learning algorithm build a predictor from known pairs of input and output where each input is a feature vector representing a log and where its corresponding output are the quantities of each lumber product produced at a given sawmill for that log. Once learned, the predictor can be used to approximate the lumbers resulting from the transformation of an unseen log at that sawmill. The 3D scans, which are made of thousands of points and which are of a variable size, were not used directly for learning.

The ICP-based prediction method we evaluate is based on the nearest neighbor algorithm [2]. The ICP method is used as a measure of distance. Based on the geometrical structure of the logs, the ICP measures the resemblance between the current log and the logs of the training set to find the closest pair. Using an ICP-based nearest neighbor, we calculate the distance between the input log and all logs from the training sets, the minimum of all distances corresponds to the log to which the input log resembles the most. One of the benefits of using ICP as a measure of distance is that it can compare two 3D scans although they might be of a different size in number of points.

### 3. THE ITERATIVE CLOSEST POINT METHOD

The ICP algorithm was introduced by Best and McKay [7] and Chen and Medioni [8] in the goal of aligning the 3D point clouds of two objects by minimizing the distance between them using geometric transformations (rotations and translations). The ICP algorithm has two steps:

- The first step consists of determining the correspondence pairs $(p, m)$ from two data sets $P$ and $X$. The goal is to find for each point $p$ in $P$ its closest point in $X$.
- The second step is to apply a transformation (rotation and translation) in order to minimize the distance between the correspondence pairs.

These two steps are repeated until the error is below a given threshold or until the maximum number of iterations is reached.

### 3.2 ICP VARIANTS

The three main different variants of the ICP found in the literature are the point-to-point method introduced by Besl and McKay [10], the point-to-plane technique by Chen and Medioni [8], and the point-to-projection method by Blais and Levine [11]. We present, in the next section, the point-to-point approach we chose for the experiment.

### 3.3 THE POINT-TO-POINT REGISTRATION METHOD

Let $X = \{\vec{x_i}\}_{1 \leq i \leq n_x}$ and $P = \{\vec{p_j}\}_{1 \leq j \leq n_p}$ be two shapes. Shape $X$ is the model shape onto which we need to align shape $P$.

Let $\vec{q_R} = [q_0\ q_1\ q_2\ q_3]^t$ be a unit quaternion vector, where $q_0 \geq 0$ and $q_0^2 + q_1^2 + q_2^2 + q_3^2 = 1$. $\vec{q_T} = [q_4\ q_5\ q_3]^t$ is the translation vector. $\vec{q} = \langle \vec{q_R} | \vec{q_T} \rangle^t$ is the transformation vector. The rotation matrix generated by $\vec{q_R}$ is given by:

$$R = R(\vec{q_R})$$

$$= \begin{bmatrix} q_0^2 + q_1^2 - (q_2^2 + q_3^2) & 2(q_1 q_2 - q_0 q_3) & 2(q_1 q_3 - q_0 q_2) \\ 2(q_1 q_2 - q_0 q_3) & q_0^2 + q_2^2 - (q_1^2 + q_3^2) & 2(q_2 q_3 - q_0 q_1) \\ 2(q_1 q_3 - q_0 q_2) & 2(q_2 q_3 - q_0 q_1) & q_0^2 + q_3^2 - (q_1^2 + q_2^2) \end{bmatrix} \quad (1)$$

Let $\vec{p} = (p_1, p_2, p_3)$ be a point of $P$ and $\vec{x} = (x_1, x_2, x_3)$ be a point of $X$. The Euclidean distance between $\vec{p}$ and $\vec{x}$ is:

$$d(\vec{p}, \vec{x}) = \|\vec{p} - \vec{x}\| = \sqrt{(x_1 - p_1)^2 + (x_2 - p_2)^2 + (x_3 - p_3)^2} \quad (2)$$

The distance between a point $\vec{p}$ and the model shape $X$ is:

$$d(\vec{p}, X) = \min_{1 \leq i \leq n_x} d(\vec{p}, \vec{x_i}) \quad (3)$$

The distance between $X$ and the closest point $\vec{x_j}$ of $X$ is defined by:

$$d(\vec{p}, \vec{x_j}) = d(\vec{p}, X) = \min_{1 \leq i \leq n_x} \|\vec{x_i} - \vec{p}\| \quad (4)$$

For $X = \{\vec{x_i}\}_{1 \leq i \leq n_x}$ and $P = \{\vec{p_j}\}_{1 \leq j \leq n_p}$ with $n_x = n_p$, the error function to be minimized is given by:

$$f(R, T) = \frac{1}{n_p} \sum_{i=1}^{n_p} \|\vec{x_i} - (R\vec{p_i} + \vec{T})\|^2 \quad (5)$$

The goal of this algorithm is to find the optimal transformation $(R, T)$ that minimizes $f(R, T)$.

The center of mass of $P$ and $X$ are given by:

$$\vec{\mu_p} = \frac{1}{n_p}\sum_{i=1}^{n_p} \vec{p_i} \qquad (6)$$

$$\vec{\mu_x} = \frac{1}{n_x}\sum_{i=1}^{n_x} \vec{x_i} \qquad (7)$$

The cross-covariance matrix of *P* and *X* is :

$$\Sigma_{px} = \frac{1}{n_p}\sum_{i=1}^{n_p}[(\vec{p_i} - \vec{\mu_p})(\vec{x_i} - \vec{\mu_x})^t] = \frac{1}{n_p}\sum_{i=1}^{n_p}[\vec{p_i}\,\vec{x_i}^t] - \vec{\mu_p}\,\vec{\mu_x}^t \qquad (8)$$

In order to calculate the vector $\Delta = [H_{23}\ H_{31}\ H_{12}]^t$ we use the matrix $H_{ij}$ where: $H_{ij} = (\Sigma_{px} - \Sigma_{px}^T)_{ij}$. $\Delta$ is used to calculate $Q(\Sigma_{px})$.

$$Q(\Sigma_{px}) = \begin{bmatrix} tr(\Sigma_{px}) & \Delta^t \\ \Delta & \Sigma_{px} + \Sigma_{px}{}^t - tr(\Sigma_{px})I_3 \end{bmatrix} \qquad (9)$$

Where $I_3$ is the identity matrix.

The optimal rotation vector $\vec{q_R} = [q_0\ q_1\ q_2\ q_3]^t$ corresponds to the maximum eigenvalue of $Q(\Sigma_{px})$.

The optimal translation vector is:

$$T = \vec{q_T} = \vec{\mu_x} - R\vec{\mu_p} \qquad (10)$$

The ICP algorithm is described below:

1. Given the point set *P* and the model shape *X*, the first step in the ICP algorithm is to initialize the iteration by $P_0=P$, $\vec{q_0} = [1,0,0,0,0,0,0]^t$ and $k = 0$.
2. The following steps are then applied until the convergence within a tolerance τ:
    a. Search the closest points: $Y_k = (P_k, X)$
    b. Compute the registration: $(\vec{q_k}, d_k) = Q(P_0, Y_k)$
    c. Apply the registration: $P_{k+1} = \vec{q_k}(P_0)$
    d. Stop the iteration when the change is mean-square error falls below a present threshold $\tau > 0$ specifying the desired precision of the registration: $d_k - d_{k+1} < \tau$.

ICP moves the shape object in order to best align it with the model shape. Translations and rotations are applied iteratively in order to minimize the error metric. The algorithm has been proved to converge to a local minimum.

## 4. EXPERIMENTS

The goal of the experiment is to evaluate a predictive approach based on a standard ICP algorithm at handling the 3D scans for prediction purposes. No filtering of the 3D scan is performed which means that some scans are missing points. The ICP-based predictor is compared to predictors built using the machine learning algorithms presented in [5]. The ICP algorithm used to compute the distance between the scans is a standard algorithm from the MATLAB machine learning library [7].

### 4.1 DATA

A total of 1207 logs were used for the experiments. The training set contains 724 logs (60% of the available datasets) and test set contains 483 logs (40% of the logs). As 736 of the 1207 logs only produces wood chips (they are too small to produce lumber) we also tested for a subset of our dataset from which we removed those (441 logs in the training set, 295 in the testing set) leading to two different datasets. We repeated the experiment 10 times, using different partitions of each dataset into a training and a test set. The sawmill has 19 lumber types leading to an output vectors $y \in \Re^{19}$.

For the ICP-based prediction method, the input for the algorithm is the 3D vector containing the points that represent the log on a 3D axis. This vector is variable in length since the number of points in each 3D scans varies.

### 4.2 PERFORMANCE EVALUATION

We evaluate the performance of the predictors built by each algorithm, including the ICP-based algorithm, with the metrics presented using the following metrics [5]: the zero-one loss, the hamming distance, the augmented hamming distance, the prediction score, the production score, and the prediction and production area score.

A *zero-one score* $s^z$ of 1 indicates that the predicted output basket $\hat{y} \in N^p$ equals the real output basket $y \in N^p$, otherwise $s^z$ is equal to 0 [9]:

$$s^z(\hat{y}, y) = \begin{cases} 1 & if\ \hat{y} = y \\ 0 & otherwise \end{cases} \tag{11}$$

Given an output vector $y \in N^p$ and a predicted output $\hat{y} \in N^p$, the *hamming distance* $d^H$ represents the average of number of prediction errors across products [9]:

$$d^H(\hat{y}, y) = {1}/{p} \sum_{j=1}^{p} f(\hat{y}_j, y_j) \tag{12}$$

where:

$$f(\hat{y}_j, y_j) = \begin{cases} 1 & if\ \hat{y}_j \neq y_j \\ 0 & otherwise \end{cases} \tag{13}$$

Given an output basket $y \in N^p$ and a predicted output $\hat{y} \in N^p$, the *augmented hamming distance* $d^{H+}$ is the sum of the ratios of the minimum between the predicted and the real quantity over the maximum of these quantities averaged across products [5].

$$d^{H+}(\hat{y}, y) = {1}/{p} \sum_{j=1}^{p} \left(1 - f(\hat{y}_j, y_j)\right) \tag{14}$$

where:

$$f(\hat{y}_j, y_j) = \begin{cases} 1 & if\ \hat{y}_j = y_j \\ \frac{\min(\hat{y}_j, y_j)}{\max(\hat{y}_j, y_j)} & otherwise \end{cases} \tag{15}$$

Given an output basket $y \in N^p$ exists also in the predicted output $\hat{y} \in N^p$, the *prediction ratio score* $s^{pre}$ is the average bounded ratio of the real production on the predicted production [5].

$$s^{pre}(\hat{y}, y) = {1}/{p} \sum_{j=1}^{p} \min\left(1, \frac{\max(\hat{y}_j, \varepsilon)}{\max(y_j, \varepsilon)}\right) \tag{16}$$

Given an output basket $y \in N^p$ exists also in the predicted output $\hat{y} \in N^p$, the *production ratio score* $s^{pro}$ is the average bounded ratio of the predicted production on the real production [5]:

$$s^{pro}(\hat{y}, y) = {1}/{p} \sum_{j=1}^{p} \min\left(1, \frac{\max(y_{j_j}, \varepsilon)}{\max(\hat{y}_{j_j}, \varepsilon)}\right) \tag{17}$$

Given an output basket $y \in N^p$ exists also in the predicted output $\hat{y} \in N^p$, the *production and prediction area score* $s^{pro\ x\ pre}$ is computed as the multiplication of the production and the prediction scores [5]:

$$s^{pro\ x\ pre}(\hat{y}, y) = s^{pro}(\hat{y}, y)\ s^{pre}(\hat{y}, y) \tag{18}$$

The ground-truth output $y$ and the predicted output $\hat{y}$ are filtered before computing the scores to avoid overestimating the predictive performance of the evaluated predictor. The filtering consists in removing the product pairs for which both the real and the predicted values are 0.

### 4.3 RESULTS AND DISCUSION

Using the ICP algorithm, it was possible to determine for each input log from the test set, the training log to which it resembles the most. As a final phase the performance scores were calculated. Table 1 contains the average scores for the ICP-based predictor as well as the average scores obtained by the predictors built using machine learning algorithms (as presented in [5]).

*Table 1: Average test scores on 10 runs with random partitions of the data*

| Scores | Without empty baskets | | | | | | |
|---|---|---|---|---|---|---|---|
| | MEAN | DT | RF | KRR | KRR-NO | K-NN | ICP |
| $s^z$ | .0722 | .5834 | .6088 | .5376 | .5519 | .56 | .1705 |
| $1-d^H$ | .1298 | .6051 | .6331 | .5809 | .5917 | .5874 | .1879 |
| $1-d^{H+}$ | .2037 | .7172 | .7423 | .7046 | .7118 | .6989 | .2662 |
| $s^{pre}$ | .537 | .841 | .8538 | .8402 | .8432 | .7982 | .5988 |
| $s^{pro}$ | .6666 | .8762 | .8886 | .8644 | .8686 | .9007 | .6674 |
| $s^{pro \, x \, pre}$ | .3109 | .7571 | .7779 | .7426 | .7511 | .7344 | .4071 |

| Scores | With empty baskets | | | | | | |
|---|---|---|---|---|---|---|---|
| | MEAN | DT | RF | KRR | KRR-NO | K-NN | ICP |
| $s^z$ | .1905 | .7006 | .7265 | .6841 | .6919 | .6979 | .3805 |
| $1-d^H$ | .2102 | .7159 | .7398 | .7077 | .7138 | .7137 | .4037 |
| $1-d^{H+}$ | .2428 | .7839 | 8044 | .783 | .786 | .7813 | .4798 |
| $s^{pre}$ | .6332 | .8894 | .8945 | .8885 | .8904 | .8635 | .7775 |
| $s^{pro}$ | .6096 | .8944 | .9099 | .8946 | .8957 | .9178 | .7022 |
| $s^{pro \, x \, pre}$ | .2964 | .8056 | .8256 | .8044 | .8077 | .8012 | .5427 |

It is worth mentioning that without the filtering of the couples equal to zero during the computation of the scores (see Section 4.2), the ICP-based predictor reached more than 90% accuracy with the $s^{pro \, x \, pre}$ score.

When filtering the real and the predicted baskets before computing the scores, the evaluation scores for the ICP-based predictor inevitable goes down. This behavior was also observed for the standard machine learning predictors [5]. In all cases, the scores of the ICP-based predictor are better than the MEAN algorithm while being worse than the ones obtained by the predictor built using a machine learning approach. This is expected since the ICP-based predictor uses a simple nearest neighbor approach. These results also highlight some of the difficulties encountered by an ICP-based predictor while comparing the 3D scans. We recall that no pre-filtering of the scans is made in this case.

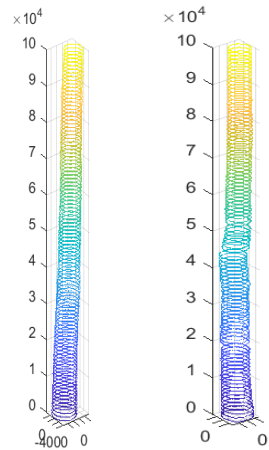

*Figure 3: 3D representations of two logs for which the ICP distance is large*

The first encountered difficulty is that some logs, although separated by a large ICP distance, share the same basket of products. This is the case of the logs presented in Figure 3. These logs are separated by a large distance using the ICP algorithm, and they are not considered as similar, even though they have the same basket of products. Our hypothesis is that more data will help in overcoming this issue.

Another difficulty encountered is that some logs that are similar do not share the same basket of products. This is the case for the in the example of Figure 4. The two logs have a high level of similarity according to the ICP distance, but the first log has an empty basket of products and the second one has a basket that contains two products type 2.

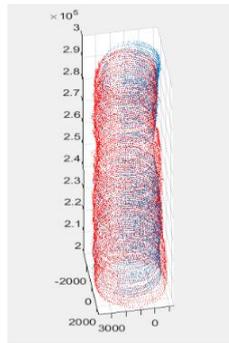

*Figure 4: 3D representations of two logs for which the ICP distance is small*

### 5. CONCLUSION

In this paper, we used the ICP algorithm to compute the distance between pairs of 3D log scans. By minimizing the distance between two given logs, this method is able to determine to which log from a training set an unseen log resembles the most although the 3D scans might differ in the number of points.

The ICP associates a distance to the compared pairs, when this distance is too large, we can understand that the compared couple doesn't have a high level of similarities. This intuition leads to the evaluation of a simple predictor based on the nearest neighbor algorithm.

Coupling the ICP with a machine learning method is one of the perspectives to be considered for the amelioration of the prediction process. As also discussed in the experiments section, we saw that direct use of the logs' 3D scans by an ICP-based predictor might require specific normalization procedures.